\documentclass{article}
\usepackage{graphicx}
\usepackage{amssymb,amsmath}
\usepackage{multicol}
\usepackage{eqnarray}
\usepackage{caption}
\usepackage{subcaption}
\usepackage{textcomp}
\usepackage{multirow}
\usepackage{float}
\usepackage{bm}
\usepackage{color}
\usepackage{epstopdf}
\usepackage{hyperref}
\usepackage{geometry}
\usepackage[linesnumbered,ruled]{algorithm2e}
\begin{document}
\title{A High GOPs/Slice Time Series Classifier for Portable and Embedded Biomedical Applications} 
\date{} 
\author{Hamid Soleimani, Aliasghar, Makhlooghpour, Wilten Nicola, \\Claudia Clopath, Emmanuel.~M.~Drakakis}
\maketitle{} 
\begin{abstract}
Nowadays a diverse range of physiological data can be captured continuously for various applications in particular wellbeing and healthcare. Such data require efficient methods for classification and analysis. Deep learning algorithms have shown remarkable potential regarding such analyses, however, the use of these algorithms on low--power wearable devices is challenged by resource constraints such as area and power consumption. Most of the available on--chip deep learning processors contain complex and dense hardware architectures in order to achieve the highest possible throughput. Such a trend in hardware design may not be efficient in applications where on--node computation is required and the focus is more on the area and power efficiency as in the case of portable and embedded biomedical devices. This paper presents an efficient time-series classifier capable of automatically detecting effective features and classifying the input signals in real--time. In the proposed classifier, throughput is traded off with hardware complexity and cost using resource sharing techniques. A Convolutional Neural Network (CNN) is employed to extract input features and then a Long-Short-Term-Memory (LSTM) architecture with ternary weight precision classifies the input signals according to the extracted features. Hardware implementation on a Xilinx FPGA confirm that the proposed hardware can accurately classify multiple complex biomedical time series data with low area and power consumption and outperform all previously presented state--of--the--art records. Most notably, our classifier reaches 1.3$\times$ higher GOPs/Slice than similar state of the art FPGA--based accelerators.
\end{abstract}

\section{Introduction}
Advent of technologies such as wearable sensor systems could be an answer to the rising issues such as increasing individuals with critical medical conditions, providing quality care for remote areas and methods to maximize the participation of disable patients \cite{patel} that healthcare system struggle with. Chronicle electronic health data that can reformed to the time series in machine learning tasks are prominent information should be sensed and analyzed using human biologically activities \cite{mazilu}. The interest for wearable systems originates from the need for monitoring patients over extensive periods of time \cite{Subhas}. Wearable activity systems mainly include sensors such as accelerometers, gyroscopes or magnetic field communication/chemical sensors \cite{bulling}, communication systems and process systems for analyzing generated signals. Smart wearable sensors are effective and reliable for preventative methods in many different facets of medicine such as, cardiopulmonary, vascular \cite{Geoff}. Further, the use of wearable sensors has made it possible to have the necessary treatment at home for patients after heart-attacks and diagnosis of some heart diseases such as cardiac \cite{Subhas}.

\par Regardless of these achievements, most contemporary commercial products only can measure simple metrics such as heart beats or steps. In addition, high computational requirement to classify high dimensional, ordered attributes time series of interest makes it practically impossible in real-time. Compare to traditional time series classifiers deep learning algorithms, armed with multiple layer of feature hierarchical, capable of extracting temporal dependencies in time series and more powerful processing capacities in wearable systems pave the way for performing more data analysis on-node and in real--time. This capability to perform more complex data analysis on the wearable device/node provides the opportunity to decrease transition data from device to host, or on the other word save data bandwidth link. The bandwidth saving is more exposes itself in the heart disease patients who should continuously monitored and classified using ultrasound machines or the victims such as cardiovascular disease does not have access to health care service , even if the doctor, relatives are not near the patient and also during the non--availability of the cellular network \cite{Kala}. However, full hardware implementation of deep neural networks still challenging for designers on wearable sensors and embedded platform due to memory bandwidth and energy inefficiency of high computational units.

\par Recent studies on the development of deep learning hardware accelerators mainly have tried to achieve highest throughput, keeping up with real-time demands of complicated and embedded machine learning algorithms, led to intricate systems with a large number of Multiply Accumulate (MAC) processors. As a result, when it comes on the hardware realization regarded systems consume large silicon area ($\sim$600~mm$^2$) and power ($\sim$500~W) \cite{GPU}\cite{Google}\cite{Intel}. Based on the observation that most biological time series signals have small rates of frequency (0--500 Hz), an alternative approach, by trading off throughput and hardware complexity using sharing resources is proposed. In this paper we propose a generalized time series classifier that implements both feature extraction and sequence learning respectively through an CNN and an LSTM network. The architecture applied to multiple biomedical disease database and various tradeoffs will be investigated. It will be also shown that the proposed system is compact, portable as well as accurate. In contrast to muscle signals which can be classified using features such as amplitude of signals, heart signals classification needs to highlight more subtle features. These feature could be extracted through training an CNN. The advantage of doing so compared to other classical feature extraction methods is the ability of learning new features. The proposed generalized architecture can be reconfigurable and trained for different applications.


\begin{figure}[t]
\centering
\includegraphics[trim = 0.3in 0.1in 0.1in 0.1in, clip, width=4.0in]{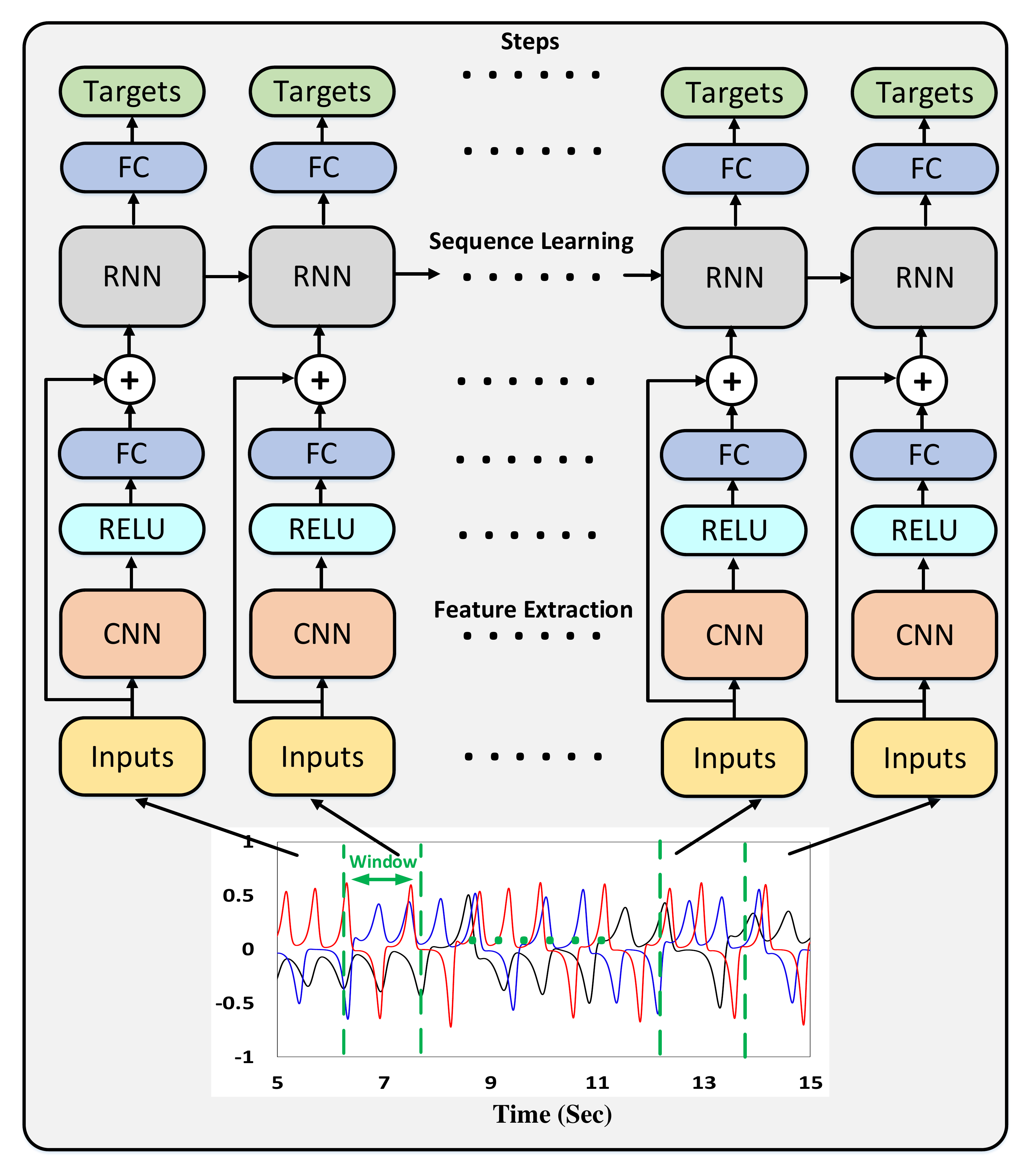}
\captionsetup{font=footnotesize}
\caption{The overall architecture of the proposed classifier along with three input time series. A portion of the input signal termed as $window$ is highlighted in green color, sequentially fed to the systems after $steps$ times.}
\vspace{-10pt}
\label{fig:arch}
\end{figure}

\section{Hardware--oriented Time--series Classifier}
The overall architecture of the proposed time series classifier is shown in Fig.~\ref{fig:arch}. In this approach, feature extraction is carried out using CNN blocks and then data is entered to the RNN blocks for sequence learning. The sequential target replication technique inspired from \cite{lipton} is used during the training phase. However our approach is slightly different. In the proposed architecture, during forward path and for all $steps$, the same output label is used to calculate the output error. The error is stored $q$ times in memory in order to calculate the gradient descent during the backward path. In the backward pass, the RNN is unrolled back in time and the weights are updated. This technique forces the network to better memorize the previous sequences of the input windows \cite{andrej}. It should be also noted that during the training phase loss values are averaged over all steps, while at the inference (test) time, the output at the final step is chosen as the actual classification value. Here is a detailed explanation of the two main blocks in the system:

\subsection{CNN Blocks}
CNNs have shown remarkable performance in image processing tasks such as object detection \cite{Ren}, face recognition \cite{Florian} and are normally composed of two types of layers: pooling layers and convolutional layers, where in this work just later layer is utilized for developing proposed architecture. Each convolutional layer is responsible for three dimension calculation of inner product of input window and weights, which referred as kernels. In contrast to regular convolution which determines output using whole input, in machine learning applications, this is done through regional products of input with a single filter. Each filter is responsible for extracting a feature from input signal. In our case, the input is a 1D time--series array, therefor the CNN filters are also 1D. The resulted output is denoted by feature map. The 1D convolution operation can be represented as follow:
\begin{equation}
\bm{z}_{(i)}^\ell = \bm{b}_{(i)}^\ell+ \sum_{a=0}^{m-1} \bm{W}_{cnn(a)}\cdot\bm{x}_{(i+a)}^{\ell - 1}
\end{equation}
where $\bm{W}_{cnn}$ and $\bm{b}$ are weight and bias of each channel, $l$ represents index of layers, $m$ is the length of each 1D filter, $\bm{x}$ and $\bm{z}$ are respectively the input and output of the network.
\par The output of each filter in an CNN layer is rectified using an activation function called ReLU which is mathematically described as follows:
\begin{equation}
\bm{r}_{(i)}^\ell = max(0,\bm{z}_{(i)}^\ell)
\end{equation}
\par The output of the activation function in the last layer is fed to a fully connected network as given here:
\begin{equation}
\bm{P}_{(i)} = \sum_{c=1}^{f-1} \sum_{d=0}^{n-1} \bm{W}_{f(c)}\cdot\bm{r}_{(d)}
\end{equation}
where $\bm{W}_{f}$ is weight of the full connection layer, $f$ is the number of filters per layer and $n$ is the length of the output feature map.
\par Current state of the art CNN networks such as ResNets \cite{Kaiming} or GoogLeNet \cite{Szegedy} have privilege of utilizing depth layers for achieving higher accuracy in the image related tasks. With increasing number of layers process of convergence becoming harder again due to the exploding/vanishing gradients. Techniques such as normalization layers \cite{Ioffe}\cite{Glorot} enabled designing networks with depth layers. In addition, these networks are vulnerable to the problem of accuracy saturations when with decreasing energy of system, accuracy does not improve \cite{Kaiming}. For dealing with this issue ResNet or GoogLeNet exploit structures such as inception module or residual learning. In our simulations we faced both problems of hampering convergence and accuracy saturations. To address these problems we used residual learning technique which is more hardware friendly and straightforward compared to the other structures. In Fig.~\ref{fig:arch} typical structure of residual module have been shown, where CNN is chosen to be two or three layers.

\subsection{RNN Blocks}
LSTM networks are very powerful Recurrent Neural Networks (RNN) that explicitly add memory gates \cite{hochreiter}. This makes the training procedure more stable and allows the model to conveniently learn both long and short--term dependencies. There are some variations on the LSTM architecture, however in this paper we use the following model \cite{schmid}:
\begin{equation}
\begin{cases}
\bm h^f_{n+1}=\sigma( \bm{W}_f^T \cdot\bm{xx}_n+\bm b_f) \\
\bm h^i_{n+1}=\sigma( \bm{W}_i^T\cdot\bm{xx}_n+\bm b_i)\\
\bm h^o_{n+1}=\sigma( \bm{W}_o^T\cdot \bm{xx}_n+\bm b_o)\\
\bm h^c_{n+1}=\tanh( \bm{W}_c^T\cdot \bm{xx}_n+\bm b_c)\\
\bm c_{n+1} = \bm{h}^f_n \circ \bm{c}_n+\bm{h}_n^c \circ \bm h\bm_n^i;\\
\bm h_{n+1}=\bm h^o_n \circ \tanh(\bm c_n);\\
\bm y_{n+1}= \bm W^T_y \bm h_n +\bm b_y
\end{cases}
\end{equation}
where $\bm{xx}_n=[\bm{h}_n,(\bm x_n+P_n)]$ and $\bm{h}_n$ and $\bm{c}_n$ are the output and cell state vectors respectively at discrete time index, $n$.  The operator $\circ$ denotes the Hadamard element by element product.  The variables $\bm h^f_n$, $\bm h^i_n$, $\bm h^o_n$ represent the forgetting, input and output gating vectors. The parameter $\bm y_n$ is a fully connected layer following the LSTM block and serves as the network output and its size is determined by the number of output classes and as well as the number of hidden neurons. Finally, $\bm W_f$, $\bm W_i$, $\bm W_o$, $\bm W_c$ ,$\bm W_y$ and $\bm b_f$, $\bm b_i$, $\bm b_o$, $\bm b_c$, $\bm b_y$ are the weights and biases for the different layers, respectively.

\par One of the main bottlenecks for the hardware realization of RNNs and convolutional neural networks (CNNs) is the large memory size and bandwidth required to fetch weights in each operation. To alleviate the need for such high bandwidth memory access, we investigate two quantization methods (binary and ternary) introduced in \cite{hubara} \cite{li} to quantize weights embedded in the network architecture. As the changes during gradient descent are small, it is important to maintain sufficient resolution otherwise no change is seen during the training process, therefore the real--valued gradients of weights are accumulated in real--valued variables. We also set the bias values to zero to achieve further efficiency in hardware realization while delivering an acceptable classification accuracy for the experimented biomedical case study. Such quantization methods can be considered as a form of regularization that can help the network to generalize.  In particular, the binary and ternary quantization are a variant of Dropout, in which weights are binarized/ternarized instead of randomly setting part of the activations to zero when computing the parameter gradients \cite{hubara}.

\par The quantization of weights in the forward path must be also reflected in the calculation of the gradient descent. Here, we use the version of the straight--through estimator introduced in \cite{hubara} that takes into account the saturation effect. Consider the sign and round functions for binary and ternary quantization respectively as follows:
\begin{equation}
\begin{cases}
q_b=\text{sign}(r)\\
q_t=\text{round}(r)
\end{cases}
\end{equation}
and assume that estimators $g_{q_b}$ and $g_{q_t}$ of the gradients $\frac{\delta C}{\delta q_b}$ and $\frac{\delta C}{\delta q_t}$ are derived. Then, the straight--through estimators of $\frac{\delta C}{\delta r}$ are:
\begin{equation}
\begin{cases}
g_{r_b}=g_{q_b}1_{\mid r\mid\leq 1}\\
g_{r_t}=g_{q_t}1_{\mid r\mid\leq 1}
\end{cases}
\end{equation}
\par This implies that the gradient is applied to the weights if their real values are between 1 and -1 otherwise the gradient is cancelled when $r$ is outside the range.

\begin{figure}[t]
\vspace{-20pt}
\centering
\includegraphics[trim = 0.0in 0.0in 0.0in 0.0in, clip, width=4.2in]{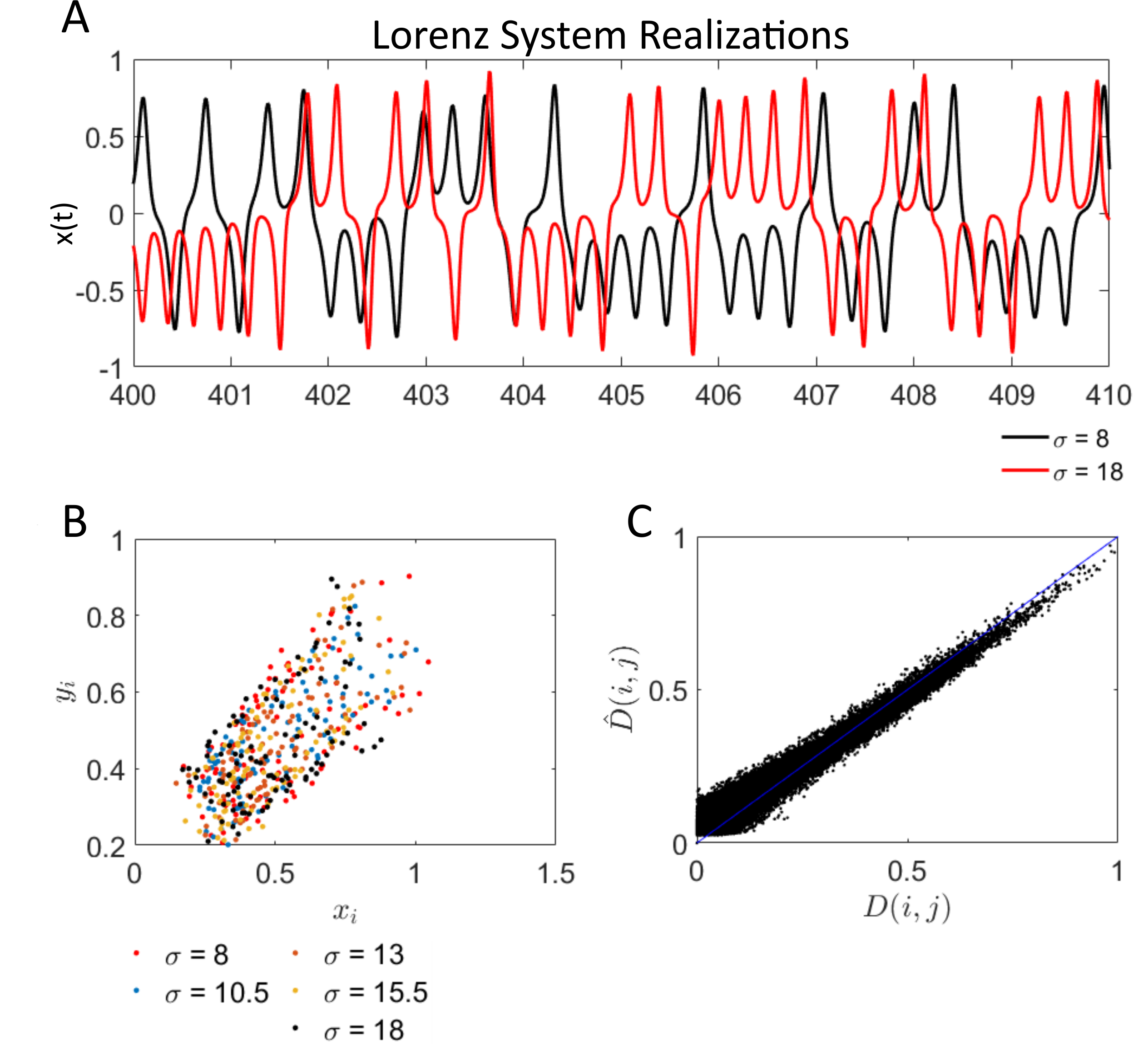}
\captionsetup{font=footnotesize}
\caption{(A) A realization of the lorenz system for $\sigma = 8$ (red) and $\sigma = 18$ (black).  The other parameters were taken to be $\rho=28$ and $\beta =\frac{5}{3}$. The $x(t)$ variable is plotted, downscaled by a factor of 40 such that it is bounded in the interval $[-1,1]$ for subsequent network training. (B) The final $(x_i,y_i)$ generated by the Fourier Spectrum and the visualization procedure outlined in the text.  The data displays no clear clustering in the Fourier domain with a 2-dimensional projection for the 5 classes considered. (C) A plot of $\hat{D}(i,j)$ vs $D(i,j)$.  The computed correlation coefficient for the Lorenz data was $ p = 0.9826$, indicating that the Fourier domain data is well described as lying on a 2-D manifold.}
\label{fig:visual}
\end{figure}

\begin{figure*}[t]
\centering
\includegraphics[trim = 0.3in 0.1in 0.3in 0.1in, clip, width=4.3in]{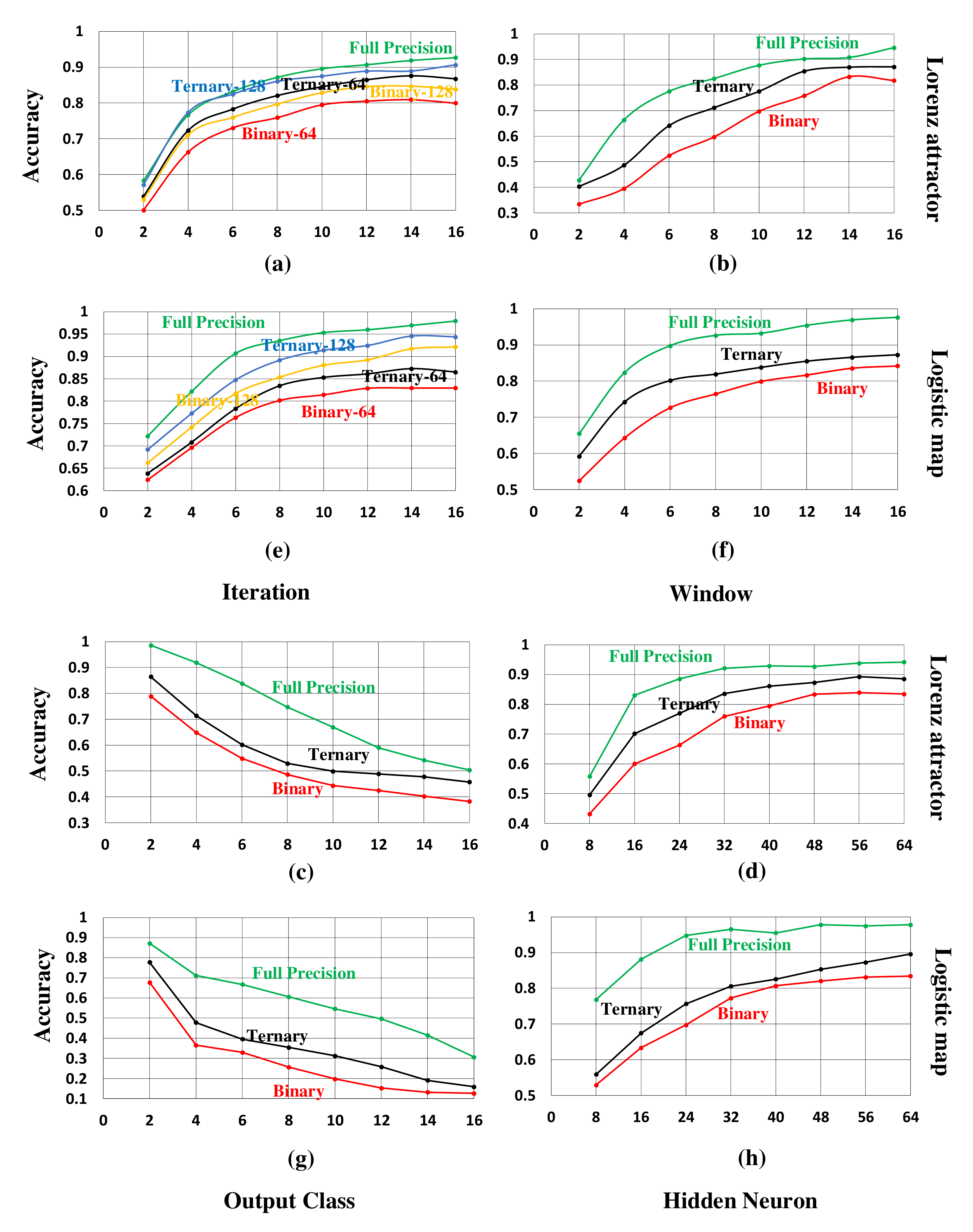}
\captionsetup{font=footnotesize}
\caption{Free parameter sweeping for three networks with various weight precision on the synthetic database. The networks' accuracies are changed by varying $iteration$ in (a) and (e), $window$ size in (b) and (f) , $output~class$ in (c) and (g) and $hidden~neuron$ in (d) and (h). In addition, two sets of binary and ternary networks are compared in (a) and (e).}
\vspace{-10pt}
\label{fig:sweep}
\end{figure*}

\section{Hardware--oriented Simulations}
In order to measure the performance of each quantization method, first we synthetically produce and classify a number of time series as a preliminary proof of principle. In this example, we generate synthetic time series data from well known chaotic dynamical systems in different parameter regimes. In all parameter regimes, these systems have chaotic attractors. The unique and isolated parameter regime corresponds to discrete classes. Only the RNN network (CNN is not included in Fig.~\ref{fig:arch} and the input time--series is directly connected to the LSTM network) is tasked to classify the resulting time series as belonging to a unique class. The considered dynamical systems are the logistic (discrete time) and Lorenz systems (continuous time). The logistic map is given by:
\begin{eqnarray*}
x_{n+1}=r\cdot x_n(1-x_n)
\end{eqnarray*}
where $x$ is the state variable and $r$ is a parameter \cite{may}.  The Lorenz system is given by:
\begin{eqnarray*}
\dot{x}&=&\sigma(y-x)\\
\dot{y}&=&x(\rho-z)-y\\
\dot{z}&=&xy-\beta z
\end{eqnarray*}
where $x$, $y$ and $z$ are state variables and $\sigma$,~$\rho$,~$\beta$ are parameters \cite{lorenz}.
\par By sweeping $r$ in the Logistic map and $\sigma$ in the Lorenz attractor, various responses can be observed (Figure 2A). The time series data generated by the chaotic dynamical systems must be similar between the classes in terms of both time and frequency features so that the signals are not easily distinguishable by the classifier.   To determine how similar the synthetic data classes are and visualize our synthetic data set in a simple way, we computed a distance matrix in the Fourier domain as
\begin{eqnarray}
D(i,j) = \int_{\omega} \left(|P_i(\omega)|-|P_j(\omega)|\right) ^2\,d\omega
\end{eqnarray}
where $P(\omega)$ is the logarithm of the power of the Fourier transforms of $x_i(t)/20$, for $i=1,2,\ldots N_S$ realizations for a sample of  $N_S=100$ realizations per discrete parameter class.  The factor of $1/20$ multiplying $x_i(t)$ bounds the trajectories in the unit interval for subsequent learning in the LSTM network. Then, we looked for a set of points in $\mathbb{R}^2$, $(x_i,y_i)$ (Figure 2B) such that $$\hat{D}(i,j) = (x_i-x_j)^2 - (y_i-y_j)^2$$ by stochastically minimizing the sum:
$$ E = \sum_{i\neq j } \left(\hat{D}(i,j) - D(i,j) \right)^2$$
This approach finds low dimensional (in this case 2D) manifolds that the data may lie on.  Alternatively, one may also use the singular value decomposition, however we do not take that approach here. The stochastic minimization occurs by initializing the $(x_i,y_i)$ from a joint uniform distribution on $[0,1]^2$, randomly perturbing every point to compute $E$. The perturbations were drawn from a normal distribution with mean 0 and standard deviation $\eta = 10^{-3}$.  At each time step, the network computes $E$ after $(x_i,y_i)$ have been perturbed and compares $E$ to the smallest value of $E$ so far, $E^*$.  If $E<E^*$, then we set the new $E^* = E$ and keep the perturbed $(x_i,y_i)$.  If $E^*<E$, we disregard the perturbation and iterate.  The results of this process are shown in Fig.~\ref{fig:visual}A-C for the Lorenz system without noise. The results demonstrate that the data has no readily visible clusters in a 2-dimensional projection, however clustering may appear in a higher dimensional projection. In order for the network to generalize the input features better, a uniform random noise is added to the training and test data. Since binarization is a form of regularization \cite{li}, we do not use other regularization methods such as Dropout. All weights are initialized by random numbers with normal distribution. The same analysis applied to the Logistic map showed that the generated time series are separable in the Fourier domain, however still difficult to visually classify in the time domain.

\par To compare the performance of the network with different weight precision, experiments with different free parameters were performed on the database. The free parameters are defined as follows:
\begin{itemize}
  \item Window~($\omega_s$):  the length of each part of the input time series fed to the network to be classified in the output. The length of input signal ($\bm u$) is equal to $M \cdot \omega_s$ where $M$ is the dimension of the input signal. For example, $M$ for data collected from a three dimensional gyroscope is 3 as the input signal is presented to the system by 3 independent time series. A highlighted window sample is shown in Fig.~\ref{fig:arch}.
  \item Iteration ($q$): the number of successive windows that must be introduced to the network sequentially so that the network can classify the input signals properly. The partition of the input signals into window sizes and then introducing them to the network sequentially would allow the RNNs to use recurrent feedback and internal memories to make decisions, leading to a significant reduction in hardware area consumption.
  \item Output~Class~($N_y$): the number of classes that the network must classify based on the input signals.  We can specify this by considering more discrete parameter sets in our chaotic systems.
  \item Hidden~neurons~($N_h$): the number of neurons embedded in the network. Accuracy increases with the number of hidden neurons at the expense of higher hardware cost.  After a certain point, there are diminishing returns in increasing the number of hidden neurons.
\end{itemize}
\par In these experiments, the LSTM network takes a sequence of continuous/discrete arrays defined by the $\omega_s$ size as input, and after $q$ steps classifies it into one of the output classes. The training objective (loss function) is the cross--entropy loss over all target sequences as follows:
\begin{equation}
L_i=-\log(p_{y_i});~p_k=\frac{e^{y_{i}}}{\sum_j e^{y_{j}}}
\end{equation}
where $k$ is the array of class scores for a single example and $p$ is a vector of the output normalised probabilities. In order to backpropagate the output error to all layers, $\frac{\delta L_i}{\delta f_{k}}$ can be derived using chain rule as follows:
\begin{equation}
\frac{\delta L_i}{\delta f_{k}}=p_k-1(y_i=k).
\end{equation}
\par Adagrad is used as the learning algorithm with learning rate of 5e-2 \cite{Duchi}. The weights are randomly drawn from a uniform distribution in [-0.01,0.01]. After each iteration, the gradients are clipped to the range [-5,5]. The results of sweeping on the free parameters of the test synthetic database for the binary, ternary and full precision networks are shown in Fig.~\ref{fig:sweep}. As can be seen in Fig.~\ref{fig:sweep} (a) and (e), by increasing $q$, the accuracy of the classifier increases at the expense of longer latency and higher power consumption for the hardware realization. It can be also observed that the quantized networks with the same number of neurons (64) can classify the input signals with a lower accuracy rate. However, this reduction in the accuracy can be compensated by increasing the number of hidden neurons. For example, 128 neurons in the quantized network have similar performance compared to 64 neurons in a full precision network. Although requiring more neurons in a quantized network,  a significant hardware efficiency improvement can be still seen. It is also observed that the ternary network possesses better accuracy performance compared to its binarized counterpart thereby confirming the results in \cite{li}.

\par The effect of varying window size $\omega_s$ on the accuracy for all networks are shown in Fig.~\ref{fig:sweep} (b) and (f). Increasing $\omega_s$, increases the accuracy for all networks but this imposes a higher hardware cost in terms of area and power. Thus, by increasing the length of the scanned input signals, the number of operations and the memory bandwidth per input increase. Moreover, these plots show that the accuracy drops as quantization is applied to the weights. Again, this reduction in accuracy can be compensated by increasing the number of hidden neurons while still achieving better hardware area performance compared to the full precision.
\par Fig.~\ref{fig:sweep} (c) and (g) show that the accuracy of the classifier drops in all networks if the number of output classes increases. However, similar to the previous experiments this reduction in accuracy can be compensated up to a certain point by increasing the number of hidden neurons as observed in Fig.~\ref{fig:sweep}~(d) and (h).

\begin{figure}[t]
\centering
\includegraphics[trim = 0.3in 0.2in 0.3in 0.2in, clip, width=3.3in]{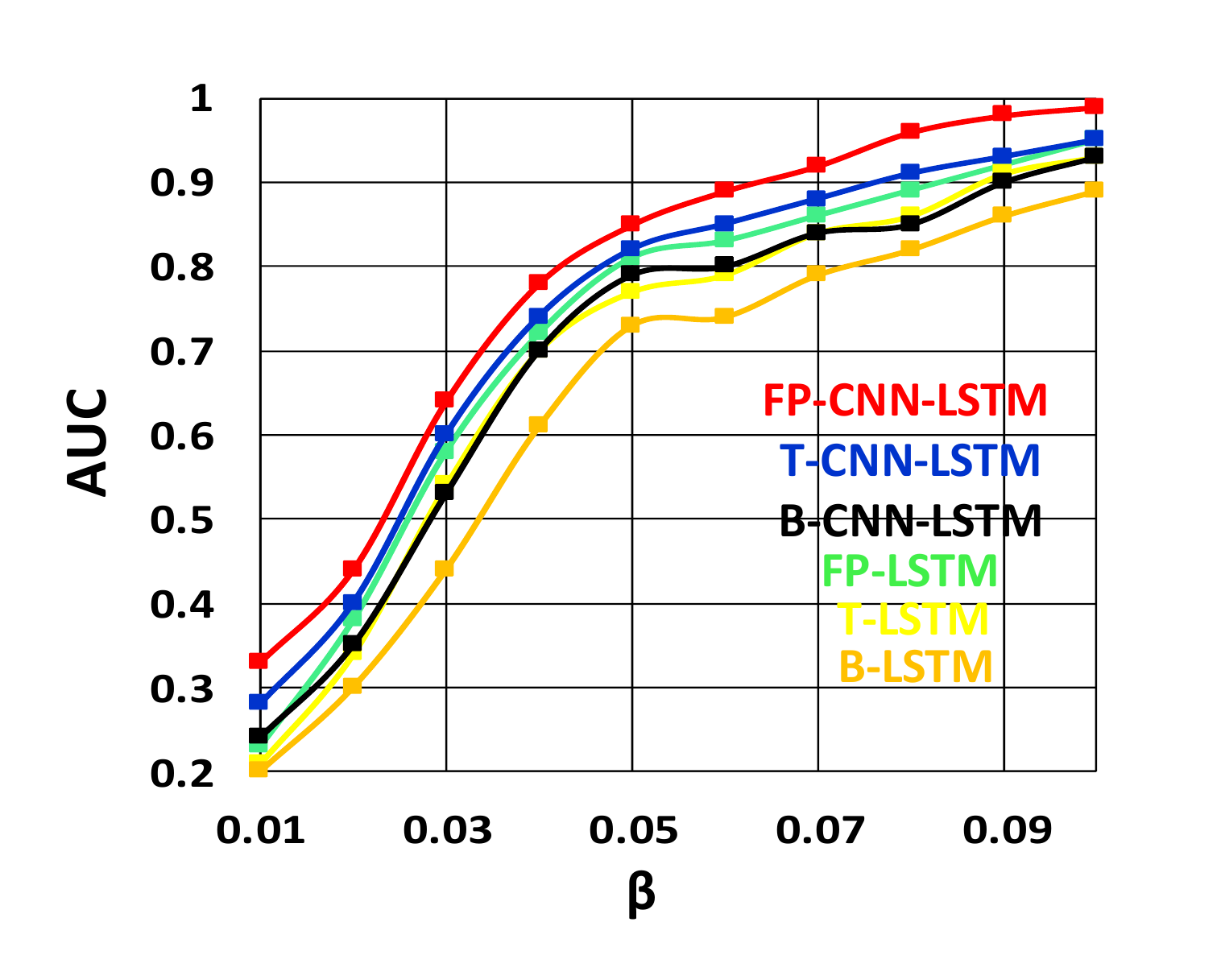}
\captionsetup{font=footnotesize}
\caption{Similarity index ($\beta$) sweeping for only LSTM network and the proposed CNN--LSTM classifier with various precision.}
\label{fig:sin}
\end{figure}

Then we investigate the impact of adding the feature extractor layer (CNN) to the classifier. We use a simple case study as follows:
\begin{equation}
\Psi_j(t)=sin(t(\alpha+j\times\beta))
\end{equation}
where $j$ is the number of output classes which is 5, $\alpha$ is a constant value which is 3 and $\beta$ is the frequency difference between each channel ranging from 0.1 to 0.01. The reason why we choose such a separable case study is that the similarity index ($\beta$) between the channels can be linearly altered. The smaller $\beta$, the higher the similarity between the classes and consequently more difficulty to differentiate the output classes. As can be seen in Fig.~\ref{fig:sin}, by reducing $\beta$, the accuracy for all networks with different precisions drops. It is also shown that the due to the added feature extractor to the network the performance of the proposed CNN-LSTM architecture is always higher than the corresponding network without CNN layer. The performance of the proposed network is also compared without CNN layer for a inseparable dynamical system's time series with certain parameter sets extracted from Lorenz attractor \cite{lorenz} as the similarity index cannot be swept due to a chaotic nature of the dynamical system. Results shown in Table~\ref{table:chaotic} confirm that the proposed CNN--LSTM architecture achieves higher performance with various precisions compared to the only LSTM classifier.

\begin{table}[t]
\captionsetup{font=footnotesize}
\caption{Classification test error rates of only LSTM networks with different weight resolutions/structure and the proposed CNN-LSTM structure trained on a inseparable dynamical system's time series. The CNN network has two layers with 20 and 50 neurons each respectively.}   
\centering          
\resizebox{\columnwidth}{!}{%
\begin{tabular}{c c c c c c}    
\hline\hline                        
  \textbf{Model} & \textbf{Learning~Rate}& \textbf{Hidden Neurons} & \textbf{Input~Window}& \textbf{Steps} &\textbf{AUC $\%$}\\ [0.5ex]  
\hline                      
\textbf{FP-LSTM} &0.05& 250 & 50& 30&90.32\\
\textbf{T-LSTM}&0.1& 350 & 50& 30 &88.37\\
\textbf{B-LSTM}& 0.1& 350 & 50 & 30& 83.23\\        
\textbf{FP-CNN-LSTM}& \textbf{0.05}& \textbf{250} & \textbf{50} & \textbf{30}& \textbf{93.43}\\        
\textbf{T-CNN-LSTM}& \textbf{0.1}& \textbf{350} & \textbf{50} & \textbf{30}& \textbf{91.65}\\        
\textbf{B-CNN-LSTM}& \textbf{0.1}& \textbf{350} & \textbf{50} & \textbf{30}& \textbf{88.91}\\ [1ex]        
\hline          
\end{tabular}
}
\label{table:chaotic}    
\end{table}

\section{Hardware Finite State Machine}
The proposed hardware classifier functions as a finite state machine that iterates through eight states and only one state is active at a time. This structure can also be implemented in a pipelined form with multiple active states and higher throughput at the expense of increased power and area consumption. The general functionality of each state is briefly described as follows:
\par $\textbf{State~1:}$ After initialization at the start state, the first input transmission according to the defined $\omega_s$ is carried out and the system enters the first state where $\sum_{a=0}^{m-1}\bm W_{cnn} x_{(i+a)}^{\ell - 1}$ is calculated with 32 additions and multiplexing per clock cycle. A total of $(\omega_s-m+1)\times f\times\frac{m}{32}$ clock pulses are needed for each layer in this state to finish the calculations as we use zero--padding and stride is equal to one. In this state after completing each CNN layer, a ReLU function ($f(x)=max(0,x)$) is also applied which adds $(\omega_s-m+1)\times\frac{f}{32}$ operations overhead.
\par $\textbf{State~2:}$ Then, according to (3) the FC layer is implemented in this state using 32 MAC operations in parallel and the result is added to $x$ as seen in Fig.~\ref{fig:arch}.
\par $\textbf{State~3:}$ The output of the previous state is formed equal to $\omega_s$ and then $\bm W_f^T \bm {xx}$, $\bm W_i^T \bm {xx}$, $\bm W_o^T \bm {xx}$ and $\bm W_c^T \bm {xx}$ are independently calculated each with eight additions per clock cycle. A total of  $(N_h+\omega_s)\times \frac{N_h}{8}$ clock pulses are needed for this state to finish the calculations.
\par $\textbf{State~4:}$ At the end of the calculations, the system enters the second state. In this state, according to (2), nonlinear functions ($\sigma(.)$ and $\tanh(.)$) are applied to the previous values fetched from memories. After $N_h$ clock pulses, the system enters the next state.
\par $\textbf{State~5:}$ In this state, using two embedded multipliers, the value of variable $\bm c$ is calculated in $N_h$ clock pulses. The critical path of the proposed architecture is limited by this state which can be alleviated by using pipelined or serial multipliers at the expense of increased latency and hardware cost.
\begin{algorithm}
    \SetKwInOut{Input}{Input}
    \SetKwInOut{Output}{Output}

    \underline{Hardware State Machine} $(x,y)$\;
    \Input{Time--series ($x$) with the length of $\omega_s$}
    \Output{Classified label ($y$)}
    \If{$state=1$}
      {
        calculate $\sum_{a=0}^{m-1}\bm W_{cnn(a)} x_{(i+a)}^{\ell - 1}$\;
        apply ReLU ($f(z)=max(0,z)$)\;
        \If{$l=2$}{
        set $state=2$\;}
        \Else
        {
        set $state=1$\;
        }
      }
      \ElseIf{$state=2$}
      {
        calculate $P_{(i)}=\sum_{c=1}^{f-1} \sum_{d=0}^{n-1} \bm{W}_{f(c)}\cdot\bm{r}_{(d)}$\;
        add $x$ to $P$\;
        set $state=3$\;}
            \ElseIf{$state=3$}
      {
      calculate $\bm W_f^T \bm {xx}$, $\bm W_i^T \bm {xx}$, $\bm W_o^T \bm {xx}$ and $\bm W_c^T \bm {xx}$\;
      set $state=4$\;}
            \ElseIf{$state=4$}
      {
      apply $\sigma(.)$ and $\tanh(.)$\;
      set $state=5$\;}
            \ElseIf{$state=5$}
      {
      calculate $\bm{h}^f_n \circ \bm{c}_n+\bm{h}_n^c \circ \bm h\bm_n^i$\;
      set $state=6$\;}
            \ElseIf{$state=6$}
      {
      apply $\tanh(.)$\;
      set $state=7$\;}
            \ElseIf{$state=7$}
      {
      calculate $\bm h^o_n \circ \tanh(\bm c_n)$\;
      set $state=8$\;}
            \ElseIf{$state=8$}
      {
      classify the output label ($\bm W^T \bm h_n +\bm b_y$)\;
      set $state=1$\;
      return $y$\;}

    \caption{Finite state machine algorithm for the proposed hardware classifier.}
\end{algorithm}

\par $\textbf{State~6:}$ As the variable $c$ is provided, the system enters this state where the $\tanh(.)$ function is applied to the previous values, taking $N_h$ clock pulses, then the system enters the next state.
\par $\textbf{State~7:}$ In this state, using one of the two embedded multipliers, the value of variable $h$ is calculated in $N_h$ clock pulses and the system enters the next state.
\par $\textbf{State~8:}$ Finally, if the number of scan times is equal to $s$, by calculating $\bm W_y^T \bm h$ in $N_h\times N_y$ clock pulses, the system determines the classified output and exits, otherwise enters State 1. This process is repeated for each window of the input signal(s) and managed by a master controller circuit, embedded into the system.

\section{Hardware Architecture}
\par In this section, hardware description of the proposed classifier is presented. The main goal of the design is to exploit the slow nature of physiological signals in particular heart activities for reducing hardware complexity and cost. In principle, in such systems, given that the classification rate is low, a few calculations per high speed clock (100 Mhz) are enough to handle the computational burden of the classifier. This architecture would also allow us to actively and efficiently reconfigure the system according to the user's specifications for the set number of neurons, window size, iterations and input/output classes.

\par
\begin{figure}[t]
\vspace{-10pt}
\centering
\includegraphics[trim = 0.2in 0.1in 0.2in 0.1in, clip, width=4.0in]{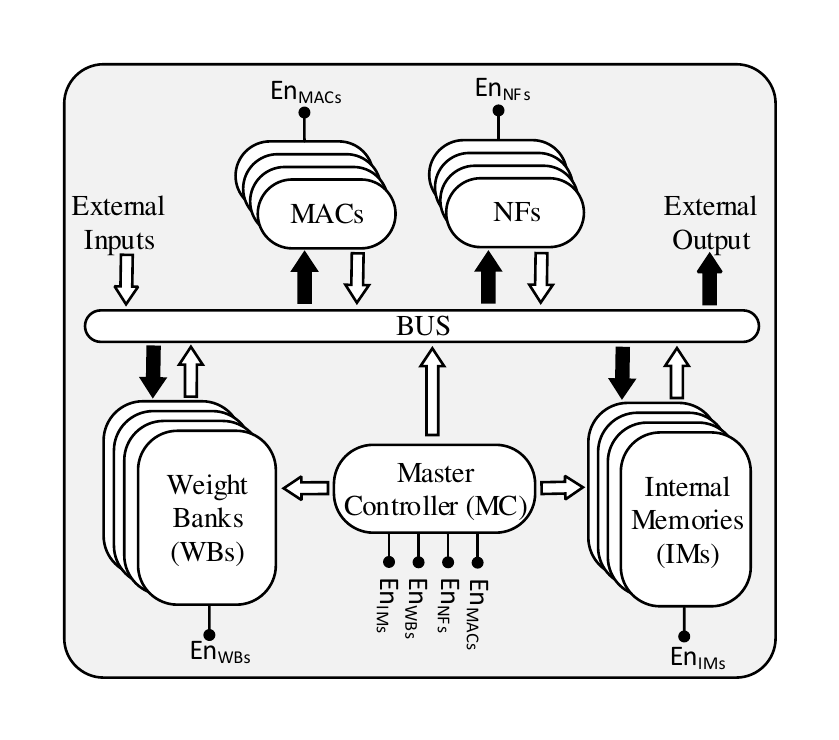}
\vspace{-10pt}
\captionsetup{font=footnotesize}
\caption{The proposed architecture of the system comprising six blocks: MAC1s, NFs, MAC2s, WBs, IMs and MC. Black and white arrows represent output and input signals respectively in accordance with bus connections}
\vspace{-10pt}
\label{fig:detail}
\end{figure}

 Other design strategies such as implementing a large number of MAC units in hardware do not apply here as high throughput is not demanded. The architecture of the proposed system is shown in Fig.~\ref{fig:detail} in which the hardware modules (maximum 64 operations per clock cycle) are shared through a 96--bit bus. It should be stressed that since the accuracy of the ternary network is generally higher than its binary counterpart as shown in Fig.~\ref{fig:sin}, in the hardware implementation, the ternary quantization is used and two bits are allocated for storing each weight value. In the following, the architecture of each hardware module is explained in detail:

\par $\textbf{WBs~(Weight~Banks):}$ This block contains five sets of buffers to store the truncated 2--bit weights ($\bm{W}_f,~\bm{W}_i,~\bm{W}_o,~\bm{W}_c$, and $\bm{W}_{cnn}$), and two sets to store full precision weights for full--connection layers ($\bm{W}_{f}$ and $\bm{W}_y$). The $WBs$ module is able to read/write maximum 64 bits in each clock cycle. The utilised volume of each buffer is defined by the user which depends on the number of hidden neurons, $f$, $\omega_s$ and etc. However, the maximum volume of these buffers must be selected based on the available resources on the FPGA. The greater the volume size, the wider the range of flexibilities for the network/input size. The reading process of this block is controlled by the $MC$ unit and the block is only activated upon its use. It should be noted that, as the proposed architecture is implemented on a Xilinx FPGA in this work, the buffers are realised using block RAMs and the address of each reading operation is provided on the negative clock edge by the $MC$ module.

\par $\textbf{IMs~(Internal~Memories):}$ This block contains sets of buffers to store 12--bit values produced by the intermediate stages for both CNN and LSTM. The maximum and utilised volume of the buffers are again determined by the available on--chip memory and the user's specifications respectively. The writing and reading of this block is also controlled by the $MC$ unit and the reading address is provided on the negative clock edge as they are implemented using block RAMs. The maximum data bandwidth of this module is 48 bits and the module is active in almost all states.
\par $\textbf{MACs~(Multply--Acummulate):}$ This blocks contains 32 parallel MAC units with full precision. Each unit computes the product of 12--bit and 2--bit or 12--bit signed numbers and adds the product to an accumulator. The number of iterations that this unit needs to operate is defined by the user and assigned by the $MC$ block.

\par $\textbf{NFs~(Nonlinear~Functions)}:$ This block is responsible for the calculation of nonlinear functions ($\sigma(u)$ and $\tanh(u)$) and ReLU employed in the hardware state machine where $u$ is the input value of these functions. We store $N$ discrete values of each nonlinear function in look-up tables with 10-bit length. The quantity $u[i]$ is the $i$th stored value where $ i\in \textbf{N}\equiv {0,1,...,N-1}$. The address of each stored value is defined as:
\begin{equation}
U[n]= \lfloor \frac{u-u_{min}}{\Delta u} \rfloor ~~~;~~~~u \in [u_{min},u_{max})
\end{equation}
where $\Delta u= \frac{u_{max}-u_{min}}{N}$. If the $u_{max}$ and $u_{min}$ parameters are orders of 2, the division in (11) can be easily performed by arithmetic shifts. Therefore, the values of these parameters are respectively chosen to be 8 and -8 for $\sigma(.)$ function and 4 and -4 for $\tanh(.)$ function. According to (11), by preparing the address, the corresponding output value can be fetched in one clock pulse from the look-up table. In our experimental setup, the value of $N$ is considered to be 64, providing enough accuracy for the calculation of nonlinear functions.

\par $\textbf{MC~(Master~Controller)}:$ According to the defined parameters by the user, this block manages and controls all resources used in the architecture through a shared bus and controlling signals ($En_{MACs}$, $En_{NFs}$, $En_{IMs}$ and $En_{WBs}$). In other words, this block actively changes the state of the FSM by assigning proper tasks to the hardware modules and actively turning off the unused modules.

\begin{table}[t]
\captionsetup{font=footnotesize}
\caption{Performance comparison between the proposed system, implemented the Kintex--7 (XC7K325T) FPGA and other state of the art hardware.}   
\centering          
\resizebox{\columnwidth}{!}{%
\begin{tabular}{c c c c c c}    
\hline\hline                        
  & \textbf{2015\cite{sankaradas}} & \textbf{2016\cite{motamedi}} &\textbf{2016\cite{qiu}}&\textbf{2017\cite{yonekawa}}& \textbf{This Work}\\ [0.5ex]  
\hline                      
\textbf{Precision}& 32bits float&32bits float&16bits float&binary&12bits fixed\\
\textbf{Frequency}& 100 MHz & 100 MHz&150 MHz& 150 MHz&100 MHz\\
\textbf{FPGA Chip}& VX485T& VX485T& XC7Z045&ZU9EG&XC7K325T\\
\textbf{GOPs}& 61.6& 84.2& 187.8&460.8&6.3\\
\textbf{Slice}& 75123& 75924& 52458&47950&508\\
\textbf{GOPs/Slice}& 8.12E-04& 11.09E-04& 35.8E-04&96.1E-04&124.0E-4\\[1ex]        
\hline          
\end{tabular}
}
\label{table:hardware}    
\end{table}

\section{Hardware Results}
\par To verify the validity of the proposed hardware classifier, the architecture designed in the previous section is implemented on a Genesys 2 development system, which provides a high performance Kintex--7 (XC7K325T) FPGA surrounded by a comprehensive collection of peripheral components. The device utilization for the implementation of the proposed hardware is summarized in Table~\ref{table:hardware} along with other state of the art implementations. The focus of all other implementations is mainly on the hardware realization of CNNs; however, as the nature of computations in all deep learning algorithms is the same, for the sake of comparison the implementation results of those studies are included here. The results of hardware implementations show that the proposed classifier reaches 1.3$\times$ higher GOPs/Slice than similar state of the art FPGA--based accelerators. Obviously, less power consumption is also achieved as the number of FPGA slices used in the proposed system is lower than in other state of art hardware. Such a trade off constrains the GOPs factor, which is not critical for most slow biomedical applications. The required response time of the system must be seriously considered upon such modifications. For example, by adding layers to CNN or LSTM, the amount of calculations is increased, therefore, the number of parallel MAC processors in the \textbf{MACs} module must be increased to keep the response time of the systems constant.

\begin{table}[t]
\captionsetup{font=footnotesize}
\caption{Classification test error rates of the LSTM networks with different weight resolutions/structure and the hardware results trained on the DB-a with 8 output classes.}   
\centering          
\resizebox{\columnwidth}{!}{%
\begin{tabular}{c c c c c c c}    
\hline\hline                        
  \textbf{Model} & \textbf{Learning~Rate}& \textbf{$N_h$} & \textbf{Input~Window}& \textbf{Steps}& \textbf{Loss} &\textbf{Accuracy $\%$}\\ [0.5ex]  
\hline                      
\textbf{Full precision}&0.05& 150 & 10& 15& 0.01 &97.63\\
\textbf{Full precision}&\textbf{0.05}& \textbf{150} & \textbf{5}& \textbf{30}& \textbf{0.01} &\textbf{97.73}\\
\textbf{Ternary}& 0.1& 250 & 5 & 30& 0.08 & 96.41\\        
\textbf{Hardware}& 0.1& 250 & 5 & 30& 0.09& 95.71\\ [1ex]        
\hline          
\end{tabular}
}
\label{table:accuracy_DB_a}    
\end{table}

\begin{table}[t]
\captionsetup{font=footnotesize}
\caption{Classification test error rates of the LSTM networks with different weight resolutions/structure and the hardware results trained on the DB-c with 12 output classes.}   
\centering          
\resizebox{\columnwidth}{!}{%
\begin{tabular}{c c c c c c c}    
\hline\hline                        
  \textbf{Model} & \textbf{Learning~Rate}& \textbf{$N_h$} & \textbf{Input~Window}& \textbf{Steps}& \textbf{Loss} &\textbf{Accuracy $\%$}\\ [0.5ex]  
\hline                      
\textbf{Full precision}&0.05& 250 & 15& 10& 0.06 &95.47\\
\textbf{Full precision}&\textbf{0.05}& \textbf{250} & \textbf{10} & \textbf{15}& \textbf{0.04} &\textbf{95.85}\\
\textbf{Ternary}& 0.1& 350 & 10 & 15& 0.12 & 94.52\\        
\textbf{Hardware}& 0.1& 350 & 10 & 15& 0.14& 93.83\\ [1ex]        
\hline          
\end{tabular}
}
\label{table:accuracy_DB_c}    
\end{table}

\begin{figure}[t]
\centering
\includegraphics[trim = 0.4in 0.1in 0.5in 0.1in, clip, width=5.4in]{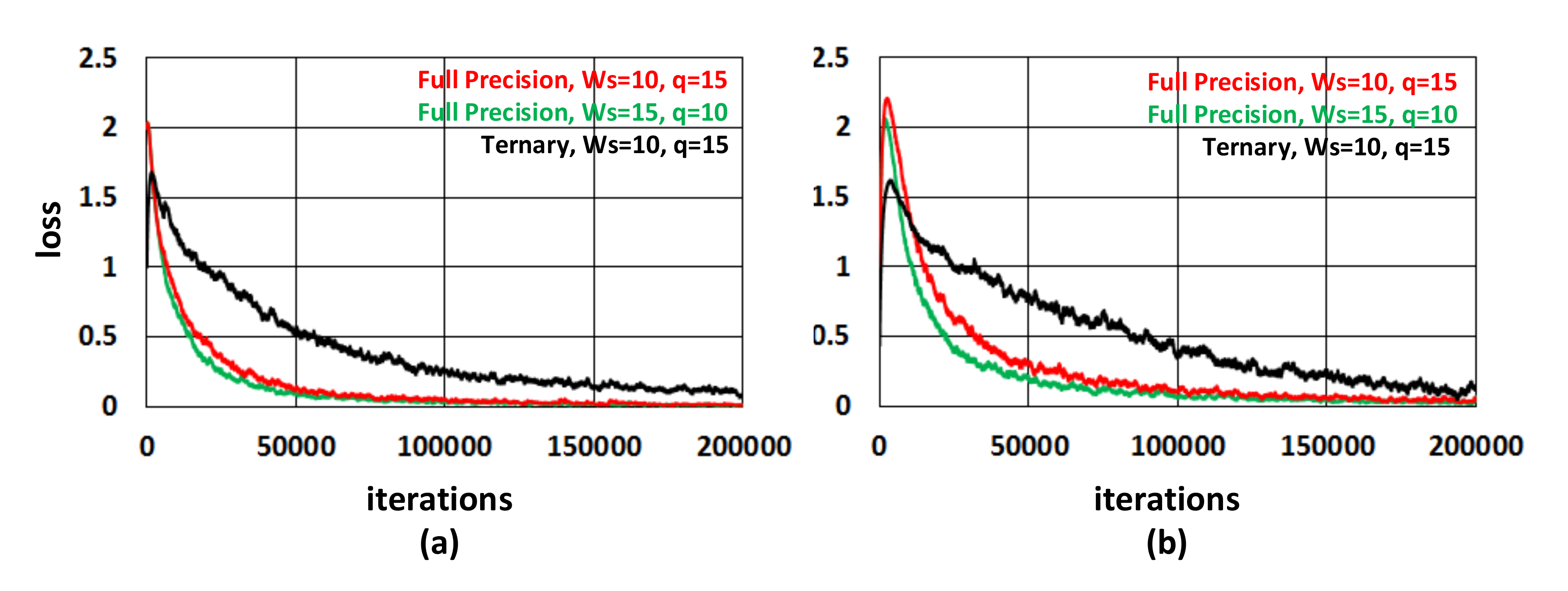}
\captionsetup{font=footnotesize}
\caption{Training loss traces for various network structures on (a) DB--a and (b) DB--c experiment. It is evident that the ternary network converges to its final value slower that full precision networks.}
\label{fig:loss}
\end{figure}

\begin{table}[t]
\captionsetup{font=footnotesize}
\caption{Confusion Matrix for DB-a database with 128 input time series and 8 output classes.}   
\centering          
\resizebox{\columnwidth}{!}{%
\begin{tabular}{c c c c c c c c c}    
\hline\hline                        
  \textbf{Output Class} & \multicolumn{8}{c}{\textbf{Target Class}}\\ [0.5ex]  
  \hline                      
 &\textbf{1}& \textbf{2} & \textbf{3}& \textbf{4}& \textbf{5} &\textbf{6}& \textbf{7} &\textbf{8} \\
\hline                      
\textbf{1}&\textbf{98.91}& 0.03 & 0.02 & 0.01& 0 &0.43& 0.01 &0.07\\
\hline                      
\textbf{2}&0.15& \textbf{96.47} & 1.02& 2.05& 0.01 &0 & 0.01 &0.27\\
\hline                      
\textbf{3}&0.14 & 0.79 & \textbf{97.80}& 0.34 & 0.53 & 0 &0.02&0.34\\
\hline                      
\textbf{4}&0& 0.57 & 0.04& \textbf{95.60} & 0.52 &0&0.26&2.98\\
\hline                      
\textbf{5}&0&0.39&0.30&2.68&\textbf{94.71} &0&0.33&1.57\\
\hline                      
\textbf{6}&0.34&0&0.02&0&0.04&\textbf{99.32} &0.26&0\\
\hline                      
\textbf{7}&0.01&0.10&0.09&0&0.02&0.36&\textbf{99.31}&0.08\\
\hline                      
\textbf{8}&0&0.07& 0.06&0.79&0.02&0.14&0.15&\textbf{98.73}\\
[1ex]        
\hline          
\end{tabular}
}
\label{table:confusion_db_a}    
\end{table}

\begin{table*}[t]
\captionsetup{font=footnotesize}
\caption{Confusion Matrix for DB-c database with 128 input time series and 12 output classes.}   
\centering          
\resizebox{\columnwidth}{!}{%
\begin{tabular}{c c c c c c c c c c c c c}    
\hline\hline                        
  \textbf{Output Class} & \multicolumn{12}{c}{\textbf{Target Class}}\\ [0.5ex]  
  \hline                      
 &\textbf{1}& \textbf{2} & \textbf{3}& \textbf{4}& \textbf{5} &\textbf{6}& \textbf{7} &\textbf{8} & \textbf{9} &\textbf{10}&\textbf{11}&\textbf{12}\\
\hline                      
\textbf{1}&\textbf{95.56}& 1.09 & 0.37& 0.96& 0.30&0.24& 0.23 &0 &0.91&0.14&0.02&0.14\\
\hline                      
\textbf{2}&0.68& \textbf{95.15} & 0.49& 0.53& 0.03&0&0.11&0.16&1.36 &0.96& 0.18 &0.24\\
\hline                      
\textbf{3}&0.66& 0.08 &\textbf{97.38}& 0& 0.41&0.85&0.30&0.18&0.08 &0&0.01&0\\
\hline                      
\textbf{4}&0.09&0 &0&\textbf{99.01}&0.26&0.09&0.21&0&0.27&0.01&0.03&0\\
\hline                      
\textbf{5}&0.01&0.26&0.31&0.44&\textbf{98.26}&0.04&0.05&0&0.02&0.36&0.19&0\\
\hline                      
\textbf{6}&0.02&0.01&0&0.39& 0 &\textbf{96.17}& 3.04 &0.01 &0.20 &0&0.11 &0.01\\
\hline                      
\textbf{7}&0&0.29&0&0.54& 0 &1.89&\textbf{96.37}&0&0.76&0.12&0.01 &0\\
\hline                      
\textbf{8}&0.08&0& 0.01& 0& 0 &0.35& 0.01 &\textbf{99.50}&0.03 &0.01&0 &0\\
\hline                      
\textbf{9}&0.03&0.07&0.01&0.01& 0 &0.07& 0.15 &0.32 &\textbf{97.65}&0.25& 0.53&0.14\\
\hline                      
\textbf{10}&0.09&0.18 &0&0.02& 0 &0.18&0.05 &0.01 &7.07&\textbf{89.94}& 1.21 &1.21\\
\hline                      
\textbf{11}&0.20&0.19&0.01&0.39& 0.11 &2.81& 0.79 &0.02& 2.95&0.60&\textbf{91.48}&0.41\\
\hline                      
\textbf{12}&0.37&0.51 &0.01& 0& .01 &0.07& 0.34&0.11 &0.88 &1.71& 2.47&\textbf{93.49}\\
[1ex]        
\hline          
\end{tabular}
}
\label{table:confusion_db_c}    
\end{table*}

\section{Surface Electromyography (sEMG) Case Studies}
To test the proposed architecture without the use of CNN layer, we use CapgMyo \cite{geng}, a hand gesture time--series database recorded by instantaneous surface electromyography (sEMG) as activity patterns in such signals can be detected using amplitude of signals. The data was collected by a non--invasive wearable device consisting of 8 acquisition modules. Each module contained a matrix--type (2 $\times$ 8) electrode array with an inter--electrode horizontal distance of 7.5 mm and a vertical distance of 10.05 mm. The 128 sEMG time--series were band--pass filtered at 20-380 Hz and sampled at 1,000 Hz with a 16--bit ADC conversion. Two different experiments were tested. In experiment 1, each one of 18 subjects performed 8 basic isometric and isotonic hand gestures including \textit{thumb up}, \textit{extension of index and middle, flexion of the others}, \textit{flexion of ring and little finger, extension of the others}, \textit{thumb opposing base of little finger}, \textit{abduction of all fingers}, \textit{fingers flexed together in fist}, \textit{pointing index} and \textit{adduction of extended fingers}. The result of this experiment is termed as DB--a in the database. In experiment 2, each of the 10 subjects performed 12 gestures performed the maximal voluntary contraction (MVC) force hand gestures including \textit{index flexion}, \textit{index extension}, \textit{middle flexion}, \textit{middle extension}, \textit{ring flexion}, \textit{ring extension}, \textit{little finger flexion}, \textit{little finger extension}, \textit{thumb adduction}, \textit{thumb abduction}, \textit{thumb flexion}, \textit{thumb extension}. The result of this experiment is termed as DB--c in the database.
\par First, we aim at evaluating the system by classifying the DB--a actions. Therefore, the output classes are separated into 8 different actions, and the envelope of the EMG signals (using a Hilbert Transform) are extracted and applied to the networks as inputs. The classification training loss and test accuracy rates of various networks with different sizes along with hardware results are shown in Table~\ref{table:accuracy_DB_a}. The table shows that the hardware classification rate obtained from the proposed structure is similar to the performance in \cite{geng}. It should be noted that 150 frames, (equivalent to 150 ms) is the window size suggested by several studies of pattern recognition based prosthetic control \cite{geng}. Therefore various options for $q$ and $\omega_s$ can be considered while the multiplication of both these parameters should be no more than 150. As $q\cdot\omega_s$ is the latency of the system to make the final classification decision, the hardware can be efficiently used if $\omega_s$ takes the lowest possible value while keeping $q\cdot\omega_s $ fixed by increasing $q$. In this case, the network with $\omega_s$ of 10 is chosen to be implemented on hardware. Similar experiments are performed on DB--c actions and results are reported in Table~\ref{table:accuracy_DB_c}, however the classification rate of this network is not reported in \cite{geng}. Here, again the network architecture with narrower length of input is chosen to be implemented on hardware. Results from both tables show that the proposed hardware can achieve an accuracy comparable with a full precision network with about 40$\%$ and 30$\%$ more neurons respectively for DB--a and DB--c experiments. Although the number of neurons in the quantized ternary networks is higher than in the full precision ones, a significant hardware efficiency improvement is still seen in the quantized networks. Training loss traces for both experiments with different network structures are shown in Fig.\ref{fig:loss}. Results show that the loss function in the ternary network reaches to the required minimum value, albeit slower than the full precision networks in both DB--a and DB--c experiments. Note that, this would only create delay in the training phase which is not critical as the network is trained once for every application.

\par Considering the sampling rate of 1000 Hz and according to $\omega_s$, $\frac{1}{1000}\times10=10~ms$ per input window is the required response time for the system in order to operate in real--time. According to Table. II, the required operations per input window for the DB--a experiment is $(5\times128+250)\times250=\sim220$ K operations which can be delivered in $\sim35~\mu s$ by the hardware classifier and is negligible compared to the required response time ($10~ms$). These operations may take longer for the DB--c experiment as more neurons are embedded in the network. According to Table. III, the required operations per input window for the DB--c experiment is $(10\times128+350)\times350=\sim570$ K operations which can be delivered in $\sim90~\mu s$ by the hardware classifier and again is negligible compared to the required response time ($10~ms$).

\par The confusion matrices extracted from CapgMyo dataset for DB-a and DB-c are respectively illustrated in Tables~\ref{table:confusion_db_a} and \ref{table:confusion_db_c}. In these experiments, the trained classifier is run 200000 times on DB-a and DB-c database. The confusion matrices compare target and predicted hand gesture classes during the test stage to identify the nature of the classification errors, as well as their quantities. The correct predictions for each output class are bolded in the tables. According to the similarities of the hand gestures, the tables highlight the occurring misclassifications accordingly. For example, in Table~\ref{table:confusion_db_a}, class 1 (\textit{Thumb up}) is misclassified 0.43 $\%$ as class 6 (\textit{Fingers flexed together in fist}) which is the closest gesture in the dataset compared to class 1. The same applies in Table V, where class 6 (\textit{Ring extension}) is misclassified 3.04 $\%$ as class 7 (\textit{Little finger flexion}). Fig.\ref{fig:tradeoffs} illustrates the response time of the proposed hardware classifier for various input window size and hidden neuron ($N_h$). The response time for the employed datasets (DB-a and DB-c) is shown with red square box. It should be noted that the proposed architecture can be conveniently modified for larger networks while delivering enough response time.

\section{Heart--Related Case Studies}
\par To test the proposed generalized time series classifier we conduct our simulations through three datasets related to the heart diseases extracted from well-known UCR datasets \cite{UCRArchive} and  PhysioNet 2016 and 2017 challenges \cite{Physio2017}. UCR datasets recorded heart activities by use of electrocardiography (ECG)device. Mean and variance of UCR datasets are near to zero and unit respectively. ECG5000 dataset originates  from \cite{ECG5000}, the BIDMC congestive heart failure database, consisting of records of 15 subjects, with severe congestive heart failure (NYHA class 3-4). Records of each individual recorded in 20 hours, containing two ECG signals, sampled with rate of 250 Hz, with 12 bit resolution and over range of (-10--10) mV. ECG200 was formatted at \cite{Olszewski} including two datasets, normal heartbeat and a Myocardial Infarction, the dataset is subset of \cite{Greenwald}, which contains 35 half—hour records and sampled with rate of 125 Hz. In PhysioNet 2016 \cite{Physio2016}, heart sound recordings have been collected from several contributors around the world, gathered at either a clinical or nonclinical environment, from both healthy subjects and pathological patients. The Challenge training set consists of five databases (A through E) containing a total of 3,126 heart sound recordings, lasting from 5 seconds to just over 120 seconds. All recordings have been resampled to 2,000 Hz and have been provided as .wav format. Each recording contains only one PCG lead. PhysioNet 2017 challenge data sampled and stored as 300 Hz, 16-bit A/C conversion with bandwidth (0.5--40 Hz) and (-5--5 mV) dynamic range. It should be noted 70 percent of online provided dataset allocated  for training set and the rest for testing set.  All of the datasets extracted from one channel. However, our model simply is able to handle multivariate time series by adding another dimension to the convolution layers.

\par To evaluate our algorithms four experiments are performed. In table \ref{table:Network characterestics} learning parameters along with characteristics of each network for different datasets are represented. It should be noted no preprocessing was performed on the datasets, and the CNN network automatically extract the important features from the input signal. CNN has two layers including 10 and 30 filters with the size of respectively $1\times5$ and $1\times3$. As mentioned in Section III, the size of hidden neurons is chosen to be 350 for ternary precision experiments. Simulation were performed using both Python and MATLAB and results are demonstrated in table \ref{table:Accuracy} confirming this fact that full-precision version of the proposed model outperforms all presented state-of-art records. In addition, quantized models could achieve acceptable accuracy compared with full--precision implementation and even better accuracy on some benchmarks.

\begin{table}[t]
\captionsetup{font=footnotesize}
\caption{Chosen Parameters and Characteristics of Networks for Different Dataset.}   
\centering 
\resizebox{\columnwidth}{!}{%
  \begin{tabular}{c c c c c}
    \hline\hline
    \textbf{Datasets} & \textbf{Input--Window} & \textbf{Steps} & \textbf{Hidden Size}&\textbf{Learning Rate}\\
    \hline
    \textbf{ECG200}&20&4&250&.05 \\
    \textbf{ECG5000}&20&7&250&.05\\
    \textbf{PhysioNet 2016}&50&30&250&.05 \\
    \textbf{PhysioNet 2017}&50&30&250&.05\\
    \hline
\end{tabular}
}
\label{table:Network characterestics}    
\end{table}

\begin{table}[t]
\captionsetup{font=footnotesize}
\caption{Performance Comparison of FP--CNN--LSTM, T--CNN--LSTM, Hardware Results With Other Existing Scores.}   
\centering 
\resizebox{\columnwidth}{!}{%
  \begin{tabular}{c c c c c c}
    \hline\hline
    \textbf{Datasets} & \textbf{Existing SOTA} & \textbf{FP-LSTM} &\textbf{FP-CNN-LSTM} & \textbf{T--CNN--LSTM}&\textbf{Hardware}\\
    \hline
    \textbf{ECG200} & 0.92 \cite{Wang} & 0.88& 0.96 & 0.93& 0.93\\
    \textbf{ECG5000} & 0.948 \cite{Karim} & 0.85 &0.95 & 0.93&0.92\\
    \textbf{PhysioNet 2016}&0.86 \cite{Physio2016}& 0.87& 0.90&0.86&0.85\\
    \textbf{PhysioNet 2017} & 0.86 \cite{Physio2017} &  0.81& 0.87 & 0.84& 0.83 \\
    \hline
\end{tabular}
}
\label{table:Accuracy}    
\end{table}

\begin{table*}[t]
\captionsetup{font=footnotesize}
\caption{Memory and MAC estimations for all case studies with various architecture and weight precisions.}   
\centering 
\resizebox{\columnwidth}{!}{%
  \begin{tabular}{c c c c c c c c c}
    \hline\hline
    \textbf{Datasets}  & \multicolumn{4}{c}{ \textbf{Memory(Mb)}}   & \multicolumn{4}{c}{ \textbf{MAC Operations (M)}} \\
    \hline
     & \textbf{FP-LSTM} & \textbf{T-LSTM}&\textbf{FP-CNN-LSTM} & \textbf{T--CNN--LSTM}& \textbf{FP-LSTM} & \textbf{T-LSTM}&\textbf{FP-CNN-LSTM} & \textbf{T--CNN--LSTM}\\
    \hline
    \textbf{ECG200}  & 9.07&1.03& 11.27 & \textbf{1.84}& 0.27&0.52& 0.95 & \textbf{1.21}\\
    \textbf{ECG5000}  & 9.07&1.03 &11.27 & \textbf{1.84}& 0.27&0.52& 0.95 & \textbf{1.21}\\
    \textbf{PhysioNet 2016}& 10.06&1.12& 12.26&\textbf{1.93}&0.30 &0.56& 0.98 & \textbf{1.24}\\
    \textbf{PhysioNet 2017}  & 10.06 &1.12& 12.26 & \textbf{1.93}& 0.30&0.56& 0.98 & \textbf{1.24}\\
    \hline
\end{tabular}
}
\label{table:assesment}    
\end{table*}

\par Training loss traces for all case studies with various weight precisions are shown in Fig.\ref{fig:loss}. Results show that the loss function in the ternary network reaches to the required minimum value, albeit slower than the full precision networks in both all experiments. Note that, this would only create delay in the training phase which is not critical as the network is trained once for every application.

\begin{figure}[t]
\centering
\includegraphics[trim = 0.2in 0.1in 0.2in 0.1in, clip, width=5.5in]{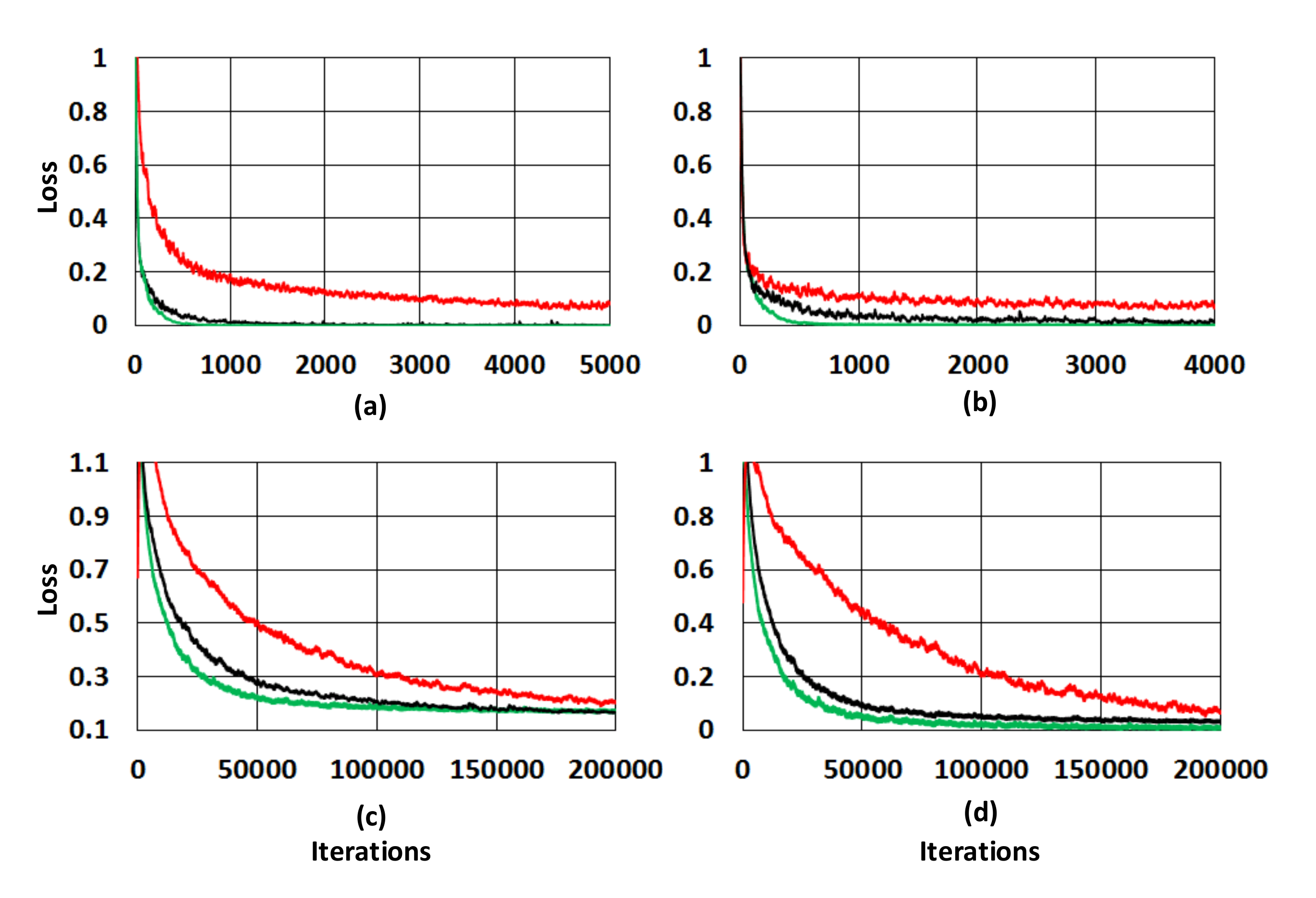}
\captionsetup{font=footnotesize}
\vspace{-15pt}
\caption{Training loss traces for various case studies (a) ECG200, (b) ECG5000, (c) PhysioNet 2016 and (d) PhysioNet 2017. FP-LSTM, FP-CNN-LSTM and T-CNN-LSTM are respectively shown with green, black and red traces. It is also evident that the ternary network converges to its final value slower that full precision networks.}
\vspace{-5pt}
\label{fig:loss}
\end{figure}

\section{Memory Assessment and MAC Operations}
\par \textbf{Memory}: There are three main sources of memory required for calculation of deep learning layers: 1) parameters including weights and biases; 2) intermediate data comes from output of each layer (e.g. features maps in CNN layers). It should be noticed that for saving memory bandwidth intermediate data is saved in on--chip memory and biases are neglected due to the minority of their sizes. Therefore, we just consider the memory required for storing weights and intermediate stages. The number of convolution weights per each layer can be estimated as follows:
\begin{eqnarray*}
\rm{CNN_{WS}} = I_d\times m\times f
\end{eqnarray*}
where $I_d$ the input depth. This value for LSTM networks equals to the following:
\begin{eqnarray*}
\rm{LSTM_{WS}}=4 \times(HN + \omega_s) \times HN
\end{eqnarray*}
where $HN$ is the number hidden neurons and $\omega_s$ is the input window size.
\par \textbf{MAC Operations}: In deep learning algorithms, MAC operation unit is normally quite dominant compared to other processing part, therefore, lower number of such units would save a significant area and latency in the design. Here, we first estimate the number of MAC operations required by the classifier, and then we will accordingly calculate the latency of the proposed hardware classifier based on the design specifications. Table \ref{table:assesment} illustrates memory and MAC estimations for for all case studies with various architecture and weight precisions. It should be stressed that the considered weight bit length for full precision (FP) is 32. As shown in the table, the required memory size for T-CNN-LSTM classifier for all case studies is lower than the FPGA Block RAM's capability implying that the model can be conveniently implemented on the FPGA. On the other hand, according to the number of required MAC operations and GOPs of the proposed hardware classifier (see Table \ref{table:hardware}), the response time is less than 0.2 mS per input window which is quite fast compared to the sampling frequency of the input heart signals.

\bibliographystyle{unsrt}	
\bibliography{paper}

\end{document}